\documentclass[sigconf]{acmart}

\copyrightyear{2026}
\acmYear{2026}
\setcopyright{cc}
\setcctype{by}
\acmConference[KDD '26]{Proceedings of the 32nd ACM SIGKDD Conference on Knowledge Discovery and Data Mining V.2}{August 09--13, 2026}{Jeju Island, Republic of Korea}
\acmBooktitle{Proceedings of the 32nd ACM SIGKDD Conference on Knowledge Discovery and Data Mining V.2 (KDD '26), August 09--13, 2026, Jeju Island, Republic of Korea}
\acmDOI{10.1145/3770855.3818061}
\acmISBN{979-8-4007-2259-2/2026/08}

\usepackage{graphicx} 
\usepackage{booktabs}
\usepackage{multirow}
\usepackage{hyperref}
\usepackage{url}
\usepackage{natbib}
\usepackage{amsfonts}
\usepackage{xspace}
\usepackage{enumitem}

\usepackage{xcolor}

\title{Temporal Graph Prototype-conditioned Conformal Prediction for Fraud Detection}

\newcommand{\ourmethod}{\textsc{ProtoCP}\xspace}
\usepackage{xcolor}
\newcommand{\Cmark}{\textcolor{green!70!black}{\Large\checkmark}}
\newcommand{\Xmark}{\textcolor{red}{\Large\texttimes}}

\author{Xudong~Chen}
\affiliation{%
  \institution{Emory University}
  \city{Atlanta}
  \country{USA}
}
\orcid{0009-0000-4892-2598}
\email{xudongchen550@gmail.com}

\author{Shengbo~Gong}
\affiliation{%
  \institution{Emory University}
  \city{Atlanta}
  \country{USA}
}
\email{shengbo.gong@emory.edu}

\author{Lu~Cheng}
\affiliation{%
  \institution{University of Illinois at Chicago}
  \city{Chicago}
  \state{IL}
  \country{USA}
}
\email{lucheng@uic.edu}

\author{Wei~Jin}
\affiliation{%
  \institution{Emory University}
  \city{Atlanta}
  \country{USA}
}
\email{wei.jin@emory.edu}

\renewcommand{\shortauthors}{Xudong Chen, Shengbo Gong, Lu Cheng, and Wei Jin}

\begin{document}

\begin{abstract}
Conformal prediction (CP) provides distribution-free coverage guarantees and has emerged as a principled tool for uncertainty quantification. In edge-level fraud detection on temporal interaction graphs, where false positives and false negatives both carry substantial cost, such coverage guarantees are particularly appealing for risk-aware decision making. However, directly applying existing graph conformal predictors yields inefficient prediction sets due to two recurring properties of fraud data. Fraudulent interactions are often embedded in benign-dominated neighborhoods that dilute calibration signals, while extreme class imbalance leaves scarce labeled-fraud support in the calibration split and leads to overly conservative class-conditional thresholds. To address these issues, we propose \ourmethod, a conformal prediction framework for edge-level fraud detection on temporal graphs. \ourmethod improves calibration efficiency by focusing calibration on fraud-relevant subgraph context and producing more stable nonconformity scores under class imbalance and temporal drift. Specifically, it leverages learned prototypes to suppress benign-dominated noise in the calibration context and introduces a neighborhood-relative scoring mechanism with temporal score diffusion for stable class-conditional calibration.
Experiments on four fraud benchmarks (YelpChi, S-FFSD, FTFD, and BankSim) show that \ourmethod achieves the target coverage with consistently smaller prediction sets than state-of-the-art baselines. Our codes are available at \url{https://github.com/Picard1701ent/ProtoCP.git}

\end{abstract}



\maketitle

\newcommand{\MainResultTable}{
\begin{table*}[t]
\caption{Comparison of methods across four fraud detection datasets in terms of Coverage and Efficiency (set size). \Cmark{} indicates that the coverage meets the target, while \Xmark{} indicates the opposite.}
\vspace{-4mm}
\centering
\resizebox{\textwidth}{!}{
\begin{tabular}{c|c|cc|cc|cc|cc}
\toprule
\multirow{2}{*}{Category} & \multirow{2}{*}{Method}
 & \multicolumn{2}{c|}{YelpChi}
 & \multicolumn{2}{c|}{S-FFSD}
 & \multicolumn{2}{c|}{FTFD}
 & \multicolumn{2}{c}{BankSim} \\
\cmidrule(lr){3-4} \cmidrule(lr){5-6} \cmidrule(lr){7-8} \cmidrule(lr){9-10}
 &  & Coverage & Efficiency($\downarrow$)
 & Coverage & Efficiency($\downarrow$)
 & Coverage & Efficiency($\downarrow$)
 & Coverage & Efficiency($\downarrow$) \\
\midrule

\multirow{3}{*}{Traditional} 
 & TPS
 & 0.89$\pm$0.00\Xmark & 1.52$\pm$0.02
 & 0.94$\pm$0.02\Xmark & 1.44$\pm$0.04
 & 0.78$\pm$0.01\Xmark & 1.51$\pm$0.01
 & 0.88$\pm$0.01\Xmark & 1.57$\pm$0.01 \\
 & APS
 & 0.91$\pm$0.01\Xmark & 1.52$\pm$0.07
 & 0.96$\pm$0.02\Cmark & 1.53$\pm$0.08
 & 0.92$\pm$0.03\Xmark & 1.48$\pm$0.09
 & 0.92$\pm$0.02\Xmark & 1.67$\pm$0.10 \\
 & RAPS
 & 0.94$\pm$0.03\Xmark & 1.48$\pm$0.08
 & 0.96$\pm$0.02\Cmark & 1.65$\pm$0.09
 & 0.93$\pm$0.03\Xmark & 1.32$\pm$0.06
 & 0.95$\pm$0.02\Cmark & 1.37$\pm$0.07 \\
\midrule

\multirow{2}{*}{Graph-based}
 & DAPS
 & 0.93$\pm$0.01\Xmark & 1.44$\pm$0.08
 & 0.94$\pm$0.02\Xmark & 1.60$\pm$0.07
 & 0.89$\pm$0.03\Xmark & 1.38$\pm$0.09
 & 0.90$\pm$0.02\Xmark & 1.34$\pm$0.08 \\
 & CF-GNN
 & 0.93$\pm$0.03\Xmark & 1.40$\pm$0.08
 & 0.94$\pm$0.03\Xmark & 1.29$\pm$0.05
 & 0.97$\pm$0.03\Cmark & 1.17$\pm$0.08
 & 0.96$\pm$0.03\Cmark & 1.13$\pm$0.06 \\

\midrule
\multirow{3}{*}{Non-exchangeable}
 & NCPNet
 & 0.99$\pm$0.02\Cmark & \textbf{1.36$\pm$0.06}
 & 0.96$\pm$0.01\Cmark & 1.30$\pm$0.04
 & 0.99$\pm$0.02\Cmark & 1.15$\pm$0.05
 & 0.99$\pm$0.02\Cmark & 1.14$\pm$0.04 \\
 & NEX
 & 0.95$\pm$0.01\Cmark & 1.56$\pm$0.08
 & 0.91$\pm$0.03\Xmark & 1.47$\pm$0.08
 & 0.96$\pm$0.03\Cmark & 1.21$\pm$0.06
 & 0.96$\pm$0.02\Cmark & 1.23$\pm$0.03 \\
 & NAPS
 & 0.95$\pm$0.02\Cmark & 1.62$\pm$0.08
 & 0.95$\pm$0.04\Cmark & 1.56$\pm$0.05
 & 1.00$\pm$0.01\Cmark & 1.16$\pm$0.04
 & 0.98$\pm$0.03\Cmark & 1.27$\pm$0.05 \\
\midrule

\multirow{3}{*}{Ours}
 & Ours
 & \textbf{0.99$\pm$0.02}\Cmark & 1.39$\pm$0.06
 & \textbf{0.97$\pm$0.02}\Cmark & \textbf{1.16$\pm$0.03}
 & 0.98$\pm$0.02\Cmark & \textbf{1.07$\pm$0.04}
 & \textbf{0.98$\pm$0.01}\Cmark & \textbf{1.04$\pm$0.03} \\
 & Ours w/o Prot
 & 0.97$\pm$0.03\Cmark & 1.48$\pm$0.07
 & 0.96$\pm$0.02\Cmark & 1.27$\pm$0.05
 & 0.96$\pm$0.02\Cmark & 1.24$\pm$0.05
 & 0.97$\pm$0.02\Cmark & 1.19$\pm$0.05 \\
 & Ours w/o Diff
 & 0.95$\pm$0.03\Cmark & 1.53$\pm$0.08
 & 0.95$\pm$0.03\Cmark & 1.19$\pm$0.05
 & 0.95$\pm$0.03\Cmark & 1.18$\pm$0.06
 & 0.96$\pm$0.02\Cmark & 1.12$\pm$0.05 \\
\bottomrule
\end{tabular}}
\label{tab:main_result}
\vspace{-3mm}

\end{table*}}

\newcommand{\SepMainTable}{
\begin{table*}[t]
\caption{Comparison of methods across four fraud detection datasets in terms of positive/negative coverage and set size. \Cmark denotes that the coverage reaches the target, while \Xmark denotes the opposite.}
\vspace{-4mm}
\centering
\resizebox{\textwidth}{!}{
\begin{tabular}{c|c|cccc|cccc|cccc|cccc}
\toprule
\multirow{3}{*}{Category}
& \multirow{3}{*}{Method}
& \multicolumn{4}{c|}{YelpChi}
& \multicolumn{4}{c|}{S-FFSD}
& \multicolumn{4}{c|}{FTFD}
& \multicolumn{4}{c}{BankSim} \\
\cmidrule(lr){3-6} \cmidrule(lr){7-10} \cmidrule(lr){11-14} \cmidrule(lr){15-18}
& 
& \multicolumn{2}{c}{Fraud}
& \multicolumn{2}{c|}{Benign}
& \multicolumn{2}{c}{Fraud}
& \multicolumn{2}{c|}{Benign}
& \multicolumn{2}{c}{Fraud}
& \multicolumn{2}{c|}{Benign}
& \multicolumn{2}{c}{Fraud}
& \multicolumn{2}{c}{Benign} \\
\cmidrule(lr){3-4}\cmidrule(lr){5-6}
\cmidrule(lr){7-8}\cmidrule(lr){9-10}
\cmidrule(lr){11-12}\cmidrule(lr){13-14}
\cmidrule(lr){15-16}\cmidrule(lr){17-18}
& & Cov. & Eff.($\downarrow$) & Cov. & Eff.($\downarrow$)
& Cov. & Eff.($\downarrow$) & Cov. & Eff.($\downarrow$)
& Cov. & Eff.($\downarrow$) & Cov. & Eff.($\downarrow$)
& Cov. & Eff.($\downarrow$) & Cov. & Eff.($\downarrow$) \\
\midrule

\multirow{3}{*}{Traditional}
& TPS
 & 0.71\Xmark & 1.68 & 0.92\Xmark & 1.50
 & 0.90\Xmark & 1.75 & 0.96\Cmark & 1.21
 & 0.73\Xmark & 1.69 & 0.78\Xmark & 1.51
 & 0.80\Xmark & 1.79 & 0.88\Xmark & 1.57 \\
& APS
 & 0.78\Xmark & 1.77 & 0.93\Xmark & 1.48
 & 0.92\Xmark & 1.79 & 0.98\Cmark & 1.33
 & 0.90\Xmark & 1.89 & 0.92\Xmark & 1.48
 & 0.83\Xmark & 1.80 & 0.92\Xmark & 1.67 \\
& RAPS
 & 0.85\Xmark & 1.83 & 0.95\Cmark & 1.43
 & 0.92\Xmark & 1.91 & 0.99\Cmark & 1.44
 & 0.85\Xmark & 1.71 & 0.93\Xmark & 1.32
 & 0.86\Xmark & 1.84 & 0.95\Cmark & 1.36 \\
\midrule

\multirow{2}{*}{Graph-based}
& DAPS
 & 0.89\Xmark & 1.73 & 0.94\Xmark & 1.40
 & 0.92\Xmark & 1.76 & 0.96\Cmark & 1.47
 & 0.81\Xmark & 1.79 & 0.90\Xmark & 1.38
 & 0.89\Xmark & 1.74 & 0.90\Xmark & 1.34 \\
& CF-GNN
 & 0.82\Xmark & 1.65 & 0.95\Cmark & 1.36
 & 0.89\Xmark & 1.47 & 0.98\Cmark & 1.15
 & 0.87\Xmark & 1.59 & 0.98\Cmark & 1.16
 & 0.90\Xmark & 1.35 & 0.96\Cmark & 1.13 \\
\midrule

\multirow{3}{*}{Non-exchangeable}
& NCPNet
 & 0.89\Xmark & 1.58 & 1.00\Cmark & 1.32
 & 0.93\Xmark & 1.49 & 0.98\Cmark & 1.16
 & 0.90\Xmark & 1.42 & 0.99\Cmark & 1.15
 & 0.89\Xmark & 1.32 & 0.99\Cmark & 1.14 \\
& NEX
 & 0.73\Xmark & 1.68 & 0.98\Cmark & 1.54
 & 0.85\Xmark & 1.66 & 0.96\Cmark & 1.32
 & 0.80\Xmark & 1.57 & 0.96\Cmark & 1.21
 & 0.83\Xmark & 1.56 & 0.96\Cmark & 1.23 \\
& NAPS
 & 0.81\Xmark & 1.74 & 0.97\Cmark & 1.60
 & 0.90\Xmark & 1.83 & 0.99\Cmark & 1.35
 & 0.91\Xmark & 1.61 & 1.00\Cmark & 1.16
 & 0.89\Xmark & 1.42 & 0.98\Cmark & 1.27 \\
\midrule

\multirow{1}{*}{Ours}
& Ours
 & \textbf{0.98\Cmark} & \textbf{1.52} & 0.99\Cmark & 1.37
 & 0.96\Cmark & 1.26 & 0.97\Cmark & 1.09
 & 0.95\Cmark & 1.36 & 0.98\Cmark & 1.07
 & 0.98\Cmark & 1.10 & 0.98\Cmark & 1.04 \\
\bottomrule
\end{tabular}}
\label{tab:sep_main_result}
\vspace{-2mm}

\end{table*}}

\newcommand{\LabelCondTable}{
\begin{table}[t]
\caption{Effect of label-conditioned calibration on S-FFSD. +LC calibrates class-specific thresholds using label-conditioned calibration scores.}
\vspace{-4mm}
\centering
\resizebox{1\columnwidth}{!}{
\begin{tabular}{c|c|cc|cc}
\toprule
\multirow{2}{*}{Method} & \multirow{2}{*}{Calib.}
& \multicolumn{2}{c|}{Fraud} & \multicolumn{2}{c}{Benign} \\
\cmidrule(lr){3-4}\cmidrule(lr){5-6}
& & Coverage & Efficiency($\downarrow$) & Coverage & Efficiency($\downarrow$)  \\
\midrule
TPS    & Global & 0.90\Xmark & 1.75 & 0.96\Cmark & 1.21 \\
TPS    & +LC    & 0.95\Cmark & 1.95 & 0.95\Cmark & 1.34  \\
APS    & Global & 0.92\Xmark & 1.79 & 0.98\Cmark & 1.33 \\
APS    & +LC    & 0.94\Xmark & 1.88 & 0.97\Cmark & 1.59 \\
CF-GNN & Global & 0.89\Xmark & 1.47 & 0.98\Cmark & 1.15 \\
CF-GNN & +LC    & 0.94\Xmark & 1.62 & 0.97\Cmark & 1.20 \\
NCPNet & Global & 0.93\Xmark & 1.49 & 0.98\Cmark & 1.16 \\
NCPNet & +LC    & 0.95\Cmark & 1.53 & 0.98\Cmark & 1.13 \\
\midrule
Ours   & Global & 0.96\Cmark & 1.26 & 0.97\Cmark & 1.09 \\
\bottomrule
\end{tabular}}
\label{tab:lc_effect}
\vspace{-4mm}

\end{table}}

\section{Introduction}

Fraud detection is critical in real-world platforms and services, including telecommunications, online review systems, and financial networks~\cite{ali2022financial, abdallah2016fraud}.
Many of these applications naturally involve graph-structured data, where entities and their interactions form complex relational patterns. As a result, Graph Neural Networks (GNNs)~\cite{wu2019comprehensive-survey,kipf2016semi,jin2021node} have been widely adopted for fraud detection tasks due to their strong ability to model relational dependencies~\cite{lin2024fraudgt, cheng2025graph, chen2024scn_gnn}. In temporal edge-level fraud detection on interaction or transaction graphs, the goal is to label each future interaction as fraudulent or benign using only historical edges. Despite their demonstrated gains in detection accuracy, existing GNN-based fraud detection methods largely overlook the equally important challenge of uncertainty quantification. Without reliable uncertainty estimates, model predictions may become overconfident, obscuring the distinction between well-supported decisions and ambiguous cases that warrant human review, which can lead to severe economic and societal consequences~\cite{wang2025evaluating}. Hence, models should offer coverage guarantees with efficient prediction sets (i.e., small sets) to enable reliable decisions and targeted human review beyond accuracy. 

Conformal prediction (CP)~\cite{shafer2008tutorial, angelopoulos2023conformal,zhou2025conformal} provides uncertainty estimates with formal reliability guarantees, generating prediction sets with finite-sample, distribution-free coverage. 
Concretely, it defines a nonconformity score measuring how poorly each example conforms to the model's prediction, and uses an empirical quantile of these scores on a held-out calibration set as the threshold. This guarantee holds under exchangeability between calibration and test examples.
Recent work has extended CP to graph learning in both static and temporal settings~\cite{wang2025non, huang2023uncertainty, song2024similarity}, demonstrating that valid coverage can be maintained under graph dependence. However, when applied to fraud detection, existing graph conformal predictors tend to produce overly large prediction sets even if validity holds.
This degradation is mainly driven by two challenges of graph-based fraud detection:

\begin{figure}
    \centering
    \includegraphics[width=1\linewidth]{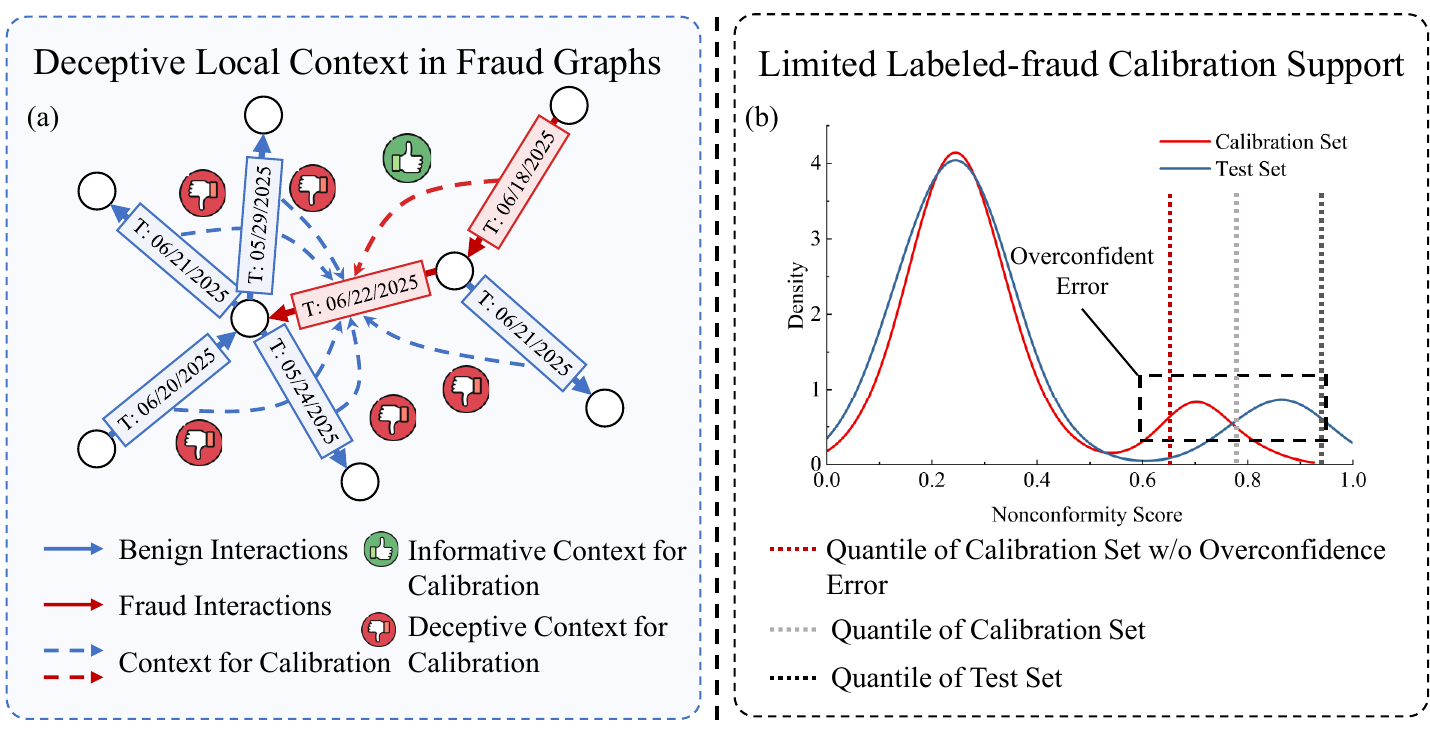}
    \vskip -1.5em
    \caption{Two graph-specific challenges for conformal prediction in fraud detection. (a) Fraudulent edges are structurally camouflaged within benign neighborhoods, which distorts neighborhood-based calibration and increases uncertainty. (b) With few fraud-labeled calibration edges, extreme nonconformity scores from overconfident errors can disproportionately raise the fraud--class calibration quantile, especially under calibration–test distribution shift.}
    \label{fig:intro}
    \vspace{-5mm}
\end{figure}

\textbf{Challenge 1: Fraud hides inside benign structure and distorts neighborhood-based calibration.} 
Fraud graphs often exhibit \textit{deceptive neighborhoods}, where the local context of a labeled fraudulent interaction is dominated by heterogeneous and frequently benign interactions (Figure~\ref{fig:intro}(a)). This dilutes fraud-relevant evidence in the neighborhood, and neighborhood-based calibration tends to become overly conservative, biasing conformal procedures toward larger prediction sets while still satisfying marginal coverage. 
For instance, in financial fraud, collusive transfers may appear only on a small subset of edges along multi-hop chains, while most surrounding interactions involve legitimate accounts~\cite{fenizia2024organized, gerbrands2022effect, dou2020enhancing, starnini2021smurf}. 
Existing methods~\cite{wang2025non, huang2023uncertainty} often overlook this fact and implicitly assume that locally similar edges exhibit similar nonconformity-score distributions, allowing them to share calibration information. Therefore, treating local neighborhoods as uniformly informative calibration context can inflate prediction sets and hurt efficiency in fraud graphs. 

\textbf{Challenge 2: Extreme rarity and temporal drift destabilize learning and calibration.} Real-world fraud datasets are often extremely imbalanced, with fraudulent cases forming only a small fraction of all interactions~\cite{dal2017credit}. This rarity directly destabilizes class-conditional CP: the fraud-class threshold must be estimated from only a handful of positive calibration samples, producing a high-variance quantile estimate. The issue is further exacerbated by overconfident misclassifications on fraud samples, which yield large nonconformity scores that act as outliers. As shown in Fig.~\ref{fig:intro}(b), when the fraud calibration sample size is small, a few extreme scores can disproportionately influence the empirical quantile, pushing up the fraud-class threshold and resulting in overly conservative prediction sets. Temporal drift further amplifies this effect, as evolving fraud patterns render test edges more difficult than calibration data, increasing the frequency of overconfident errors and magnifying the influence of extreme scores. This limitation motivates the design of stable nonconformity scores that remain robust to a small number of extreme errors, so that the fraud calibration quantile can remain less sensitive to extreme scores while maintaining efficient prediction sets at the target coverage level.

To address the above limitations, we propose \ourmethod, a fraud-aware CP framework for uncertainty quantification on fraud graphs. 
The key idea is to construct a fraud-informative calibration context for conformal scoring under coordination and camouflage, and design more stable nonconformity scores under scarce labeled-fraud calibration support. Specifically, \ourmethod addresses \textit{{Challenge 1}} by constructing a fraud-informative calibration context via prototype-guided subgraph extraction, and tackles \textit{{Challenge 2}} by stabilizing conformal calibration with prototype-conditioned nonconformity scoring under scarce labeled fraud. To be concrete, we learn class-specific prototype banks in the backbone embedding space and use an embedding-driven mask to extract a temporal subgraph that suppresses benign-dominated neighbors. In this subgraph, we augment the base nonconformity score with neighborhood-relative discrepancy features that capture prediction-level inconsistency and structural/motif deviations (e.g., degree statistics and local motif counts). Finally, to improve conformal quantile estimation under spatiotemporal dependence in fraud graphs, we apply a lightweight history-only diffusion before Mondrian calibration. To sum up, our contribution can be summarized as:
\begin{itemize}[leftmargin=*,topsep=1.5pt]
  \item  We are the first to systematically study conformal prediction for fraud detection on temporal interaction graphs, addressing challenges of benign-dominated neighborhoods and extreme rarity with temporal drifts, while explicitly targeting \emph{prediction-set efficiency} without sacrificing coverage.

  \item 
  We propose \ourmethod, a novel prototype-based conformal prediction approach that (i) learns class-specific prototypes to extract prototype-consistent temporal subgraphs that form fraud-informative calibration context, and (ii) constructs a prototype-relative nonconformity score with a lightweight history-only diffusion step before class-conditional calibration to stabilize the fraud-class quantile under scarce positives and temporal drift.

  \item 
  Experiments on multiple fraud benchmarks show that \ourmethod achieves target coverage while consistently producing tighter prediction sets than strong conformal baselines; ablations and diagnostics further verify the individual roles of prototype-con-\allowdisplaybreaks ditioned context, relative scoring, and diffusion in improving the validity-efficiency tradeoff.
\end{itemize}

\section{Related Work}

\textbf{Foundation of Conformal Prediction.}
Conformal prediction (CP)~\cite{shafer2008tutorial, angelopoulos2023conformal, vovk2005algorithmic} provides distribution-free coverage guarantees by returning prediction sets containing the true label with a user-chosen probability, under the assumption of exchangeability between calibration and test examples.
Full CP~\cite{vovk2005algorithmic} requires refitting the model for every candidate label, which is prohibitive for deep networks, while split CP~\cite{lei2018distribution} avoids this by using a held-out calibration set for nonconformity scoring.
Common score functions include TPS~\cite{sadinle2019least}, APS~\cite{romano2020classification}, and RAPS~\cite{angelopoulos2020uncertainty}, differing in how they accumulate probability mass.
When class-conditional coverage is desired, Class-conditional CP~\cite{ding2023class} calibrates within each class separately, which is particularly relevant under class imbalance.
The validity of split CP hinges on exchangeability. Weighted CP~\cite{tibshirani2019conformal} restores approximate validity under covariate shift via density-ratio reweighting, and Barber et al.~\cite{barber2023conformal} generalize to arbitrary non-exchangeable settings by bounding the coverage gap in total variation distance.
ACI~\cite{gibbs2021adaptive} further extends CP to time-series by dynamically adjusting the miscoverage rate to maintain long-run coverage under distribution shifts.
\textit{However, these methods target i.i.d.\ or sequential data and do not address the topological dependencies and temporal correlations in graph-structured fraud detection.}

\textbf{Conformal Prediction on Graphs.}
Graph-structured data challenge exchangeability, since nodes and edges are coupled through topology.
Huang et al.~\cite{huang2023uncertainty} show that exchangeability holds in the transductive setting under permutation-equivariant GNNs, while Zargarbashi~\cite{zargarbashi2024conformal} recover coverage inductively by recalculating scores upon each new arrival.
To improve set efficiency, SNAPS~\cite{song2024similarity} pools scores among structurally similar nodes, DAPS~\cite{zargarbashi2023conformal} diffuses scores over the graph, Clarkson~\cite{clarkson2023distribution} uses path-length-weighted neighborhoods as calibration context, and RR-GNN~\cite{zhang2025residual} combines graph-structured class-conditional CP with residual-adaptive scores.
Temporal graphs further introduce distribution drift that violates exchangeability even in transductive settings.
Davis et al.~\cite{davis2024valid} recover valid coverage on dynamic graphs via tensor unfolding, and NCPNet~\cite{wang2025non} extends diffusion-based calibration to temporal graphs with jointly optimized weighted quantiles.
Despite these advances, existing graph CP methods assume a reliable neighborhood-based calibration context and sufficient labeled support per class. In fraud detection graphs, both assumptions are violated: \textit{neighborhoods are deceptive and benign-dominant, diluting calibration signals, while labeled fraud examples are extremely scarce, destabilizing class-conditional thresholds. }

\textbf{Graph-based Fraud Detection.}
Graph-based fraud detection methods exploit relational structures to detect camouflage, collusion, heterophily, and abnormal interaction patterns. Representative methods improve fraud detection through relation-aware aggregation and label-balanced sampling~\cite{dou2020enhancing,liu2021pick}, spectral modeling for heterophily~\cite{wu2023splitgnn}, self-explainable mask learning~\cite{li2024sefraud}, cohort augmentation for camouflaged frauds~\cite{xiao2024vecaug}, and label-aware message passing~\cite{hyun2024lex}. These methods demonstrate the importance of graph structure for fraud detection, but they primarily optimize point-prediction accuracy and do not provide conformal prediction sets with coverage control. 
In contrast, our work focuses on uncertainty quantification for edge-level fraud detection on temporal interaction graphs, aiming to improve prediction-set efficiency while maintaining target coverage. Another related line studies graph learning with imbalanced classes, sparse labels, or distribution shifts~\cite{juan2023ins,dai2022towards, liu2022graph, guo2024investigating}, but it does not address class-wise conformal calibration under rare positives and temporal drift.

\section{Preliminary}

Conformal prediction (CP) transforms any predictive model into one that outputs prediction sets with finite-sample coverage guarantees.
We consider binary labels $Y\in\{0,1\}$, features $X\in\mathcal{X}$, and adopt the split CP framework.
Let $\mathcal{S}^{\mathrm{cal}}=\{(X^{\mathrm{cal}}_i,Y^{\mathrm{cal}}_i)\}_{i=1}^{n}$ be the calibration set and
$\mathcal{S}^{\mathrm{test}}=\{(X^{\mathrm{test}}_j,Y^{\mathrm{test}}_j)\}_{j=1}^{m}$ be the test set, with class-conditional sizes
\begin{equation}
n_y := \sum_{i=1}^{n}\mathbf{1}\{Y^{\mathrm{cal}}_i=y\}, \qquad
m_y := \sum_{j=1}^{m}\mathbf{1}\{Y^{\mathrm{test}}_j=y\}.
\end{equation}

Split CP defines a nonconformity score function $s:\mathcal{X}\times\{0,1\}\to\mathbb{R}$ that measures how poorly an example fits a candidate label, with higher values indicating greater disagreement.
After training the model on a separate training set, we compute calibration scores $S^{\mathrm{cal}}_i=s(X^{\mathrm{cal}}_i, Y^{\mathrm{cal}}_i)$ for each calibration example, and set the threshold $\hat{q}$ as the $\lceil(n+1)(1-\alpha)\rceil$-th smallest value among $\{S^{\mathrm{cal}}_i\}_{i=1}^{n}$.
The prediction set for a new test example $X^{\mathrm{test}}_j$ then includes every label whose score falls below this threshold:
\begin{equation}
\mathcal{C}(X^{\mathrm{test}}_j) := \big\{y\in\{0,1\}: s(X^{\mathrm{test}}_j, y) \le \hat{q}\big\}.
\end{equation}
Under exchangeability between calibration and test data, this guarantees marginal coverage $\Pr\big(Y^{\mathrm{test}}_j\in\mathcal{C}(X^{\mathrm{test}}_j)\big)\ge 1-\alpha$.


\section{Method}
\begin{figure*}[t]
    \centering
    \includegraphics[width=1\linewidth]{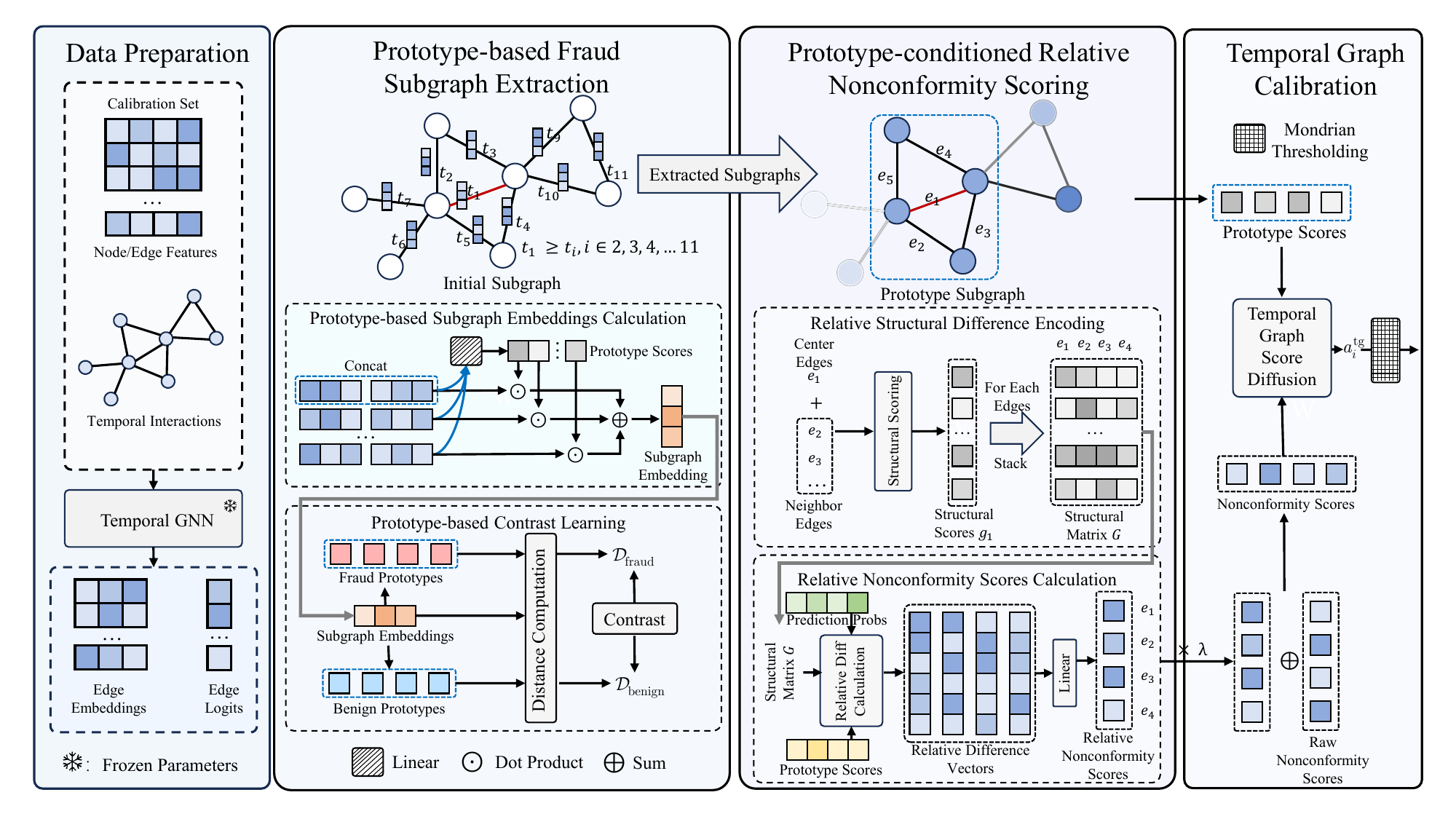}
    \vspace{-10mm}
    \caption{Overall pipeline of Prototype-based Temporal Graph Conformal Prediction}
    \label{fig:pipeline}
    \vspace{-4mm}
\end{figure*}
In this work, we propose \ourmethod, a conformal prediction framework for fraud detection on temporal transaction graphs. The key idea is to perform calibration on \emph{prototype-conditioned} subgraphs, where learned prototypes retain fraud-relevant neighbors and suppress benign noise; within these subgraphs, we construct neighborhood-relative nonconformity scores and further stabilize calibration via diffusion over the temporal graph. Accordingly, \ourmethod consists of three components: (i) \textbf{Prototype-based Fraud Subgraph Extraction}, (ii) \textbf{Prototype-conditioned Relative Nonconformity Scoring}, and (iii) \textbf{Temporal Graph Calibration}. Figure~\ref{fig:pipeline} illustrates the overall pipeline.

\subsection{Prototype-based Fraud Subgraph Extraction}\label{sec:prototype}

Fraud detection tasks often exhibit intricate group-level behavioral patterns, where fraudulent entities collaborate within highly structured relational networks (e.g., shared devices, accounts, or merchants). Existing calibration methods for conformal prediction typically treat samples independently and don't consider such relational heterogeneity. To address this issue, we introduce a prototype-based fraud subgraph extraction for the calibration stage. For each entity/transaction, we learn several prototypes to summarize common fraud-related group patterns and use them to extract representative local subgraphs for calibration. By calibrating on subgraphs that are structurally aligned with these prototypes, we reduce heterogeneity in calibration residuals and obtain more reliable uncertainty estimates under relational fraud patterns.

\textbf{Prototype Initialization. }Given a temporal graph backbone $\mathcal{F}_\theta$ (e.g., TGN~\cite{rossi2020temporal}), we encode each interaction edge $e_i=(u_i, v_i, t_i)$ into a time-dependent representation. Specifically, let $\mathcal{E}_{\le t_i}=\{(u_\ell,v_\ell,t_\ell,s_\ell): t_\ell \le t_i\}$ denote the interaction history up to $t_i$. The backbone encoder produces time-conditioned node embeddings for the endpoints of $e_i$ as
\begin{equation}
\mathbf{h}_{u_i}(t_i),\,\mathbf{h}_{v_i}(t_i)=\mathcal{F}_\theta\!\left(\mathcal{E}_{\le t_i},\,u_i,\,v_i,\,t_i\right).
\end{equation}
We then form an edge embedding 
\begin{equation}
\mathbf{x}_i=\psi\!\left(\mathbf{h}_{u_i}(t_i), \mathbf{h}_{v_i}(t_i), s_i\right),
\end{equation}
where $s_i$ denotes the edge/transaction attributes of $e_i$, and $\psi(\cdot)$ is a fusion function (e.g., concatenation), and obtain the edge-classification probability by a prediction head:
\begin{equation}
\hat{\mathbf{p}}(e_i)=\mathrm{softmax}\!\left(\mathrm{MLP}_{\text{cls}}(\mathbf{x}_i)\right),\quad
\hat{p}(y\mid e_i)=\hat{\mathbf{p}}(e_i)_y,
\label{eq:prob}
\end{equation}
where $\mathrm{MLP}_{\text{cls}}$ is the edge classifier and $\hat{p}(y\mid e_i)$ is the predicted probability for class $y$.

To characterize heterogeneous behavioral patterns across classes, we initialize a set of learnable prototypes comprising $M$ fraud prototypes $\mathcal{P}_\text{fraud} = \{\mathbf{p}_i^f\}_{i=1}^M$ and $N$ normal prototypes $\mathcal{P}_\text{norm} = \{\mathbf{p}_j^n\}_{j=1}^N$. Each prototype acts as an anchor in the edge-embedding space, summarizing a recurring interaction pattern for its class. These prototypes are subsequently refined together with the subgraph-level calibration objective, enabling pattern-aware calibration under diverse relational contexts.



\textbf{Embedding-driven Subgraph Extraction. }
For each labeled edge event $e_i=(u_i,v_i,t_i)$, we precompute its $k$-hop temporal neighborhood under a history-cutoff constraint: the subgraph is constructed only from interactions with timestamps $t \le t_i$ (i.e., up to and including the current time) in case of information leakage from future interactions.

For each $e_i$, we compute an importance weight for each neighboring edge $e_j \in \mathcal{N}_i$ by scoring the pair $(e_i,e_j)$ and normalizing over the neighborhood. Specifically, we form
\begin{equation}
\mathbf{z}_{ij}=\mathrm{concat}(\mathbf{x}_i,\mathbf{x}_j),\qquad
\alpha_{ij}=\frac{\exp\!\left(\mathrm{MLP}_{\alpha}(\mathbf{z}_{ij})\right)}{\sum_{e_k\in\mathcal{N}_i}\exp\!\left(\mathrm{MLP}_{\alpha}(\mathbf{z}_{ik})\right)},
\end{equation}
so that $\sum_{e_j\in\mathcal{N}_i}\alpha_{ij}=1$. We then obtain the subgraph representation centered at $e_i$ via attention pooling:
\begin{equation}
\mathbf{s}_i = \sum_{e_j \in \mathcal{N}_i} \alpha_{ij}\,\mathbf{x}_j.
\end{equation}


\textbf{Prototype Optimization. }To obtain semantically meaningful prototypes for prototype-aware conformal calibration, we regularize the prototype space with a margin-based objective. Let $\text{dist}(\cdot,\cdot)$ denote the cosine distance. For each subgraph embedding $\mathbf{s}_i$, we compute the nearest-prototype distances
\begin{equation}
d_i^{f}=\min_{\mathbf{p}\in\mathcal{P}_\text{fraud}} \text{dist}(\mathbf{s}_i,\mathbf{p}),\quad
d_i^{n}=\min_{\mathbf{p}\in\mathcal{P}_\text{norm}} \text{dist}(\mathbf{s}_i,\mathbf{p}).
\end{equation}
We then define a margin-based prototype loss:
\begin{equation}
\ell_i =
y_i\Big(d_i^{f} + [\gamma - d_i^{n}]_{+}\Big)
+
(1-y_i)\Big(d_i^{n} + [\gamma - d_i^{f}]_{+}\Big), \quad \mathcal{L}_\text{proto}=\frac{1}{|\mathcal{B}|}\sum_{i\in\mathcal{B}}\ell_i. 
\end{equation}
where $y_i\in\{0,1\}$ is the label (1 for fraud), $\gamma$ is the margin, and $[x]_+=\max(0,x)$. We optimize this prototype loss jointly with the conformal-prediction-related objectives during training.

\subsection{Prototype-conditioned Relative Nonconformity Scoring}
In fraud detection, structural anomalies (e.g., abnormal fan-in/out, bursty interactions, rings, or triangular motifs) are often revealed by comparing a target interaction against its local neighborhood context, consistent with prior findings on local-structure-aware graph learning~\cite{zhao2022from}. Meanwhile, under severe class imbalance, backbone models are prone to produce heavy-tailed probabilities/logits distributions for fraud class. As a result, conventional nonconformity scores based solely on absolute prediction scores (logits/probabilities) can be unreliable: they may underestimate uncertainty for minority (fraud) cases, hurting coverage, or become overly conservative, reducing calibration efficiency. By computing relative nonconformity within prototype-conditioned subgraphs, we highlight locally abnormal interactions that are not evident from an isolated edge, making neighborhood-relative deviations a more reliable resource for nonconformity than absolute confidence alone under severe class imbalance.

\textbf{Relative Nonconformity Encoding. }Given a prototype\allowbreak-condi\allowbreak tioned subgraph $\mathcal{G}_i$ centered at edge $e_i$, we obtain an importance weight $\alpha_{ij}$ for each neighboring edge $e_j\in\mathcal{N}_i$. We then construct a relative-difference feature vector by quantifying how the center edge deviates from its weighted neighborhood. Specifically, we encode three types of signals—prediction-level discrepancies, degree-based structural statistics, and motif-based counts (e.g., two-hop paths and triangles) under the same neighbor-importance weighting.

For any edge-level quantity $g(e)$ (vector or scalar), we define its weighted neighborhood mean and weighted center-neighbor absolute difference as: 
\begin{equation}
\mu_i(g)=\sum_{e_j\in\mathcal{N}_i}\alpha_{ij}\,g(e_j),\qquad
\delta_i(g)=\sum_{e_j\in\mathcal{N}_i}\alpha_{ij}\,\big|g(e_i)-g(e_j)\big|.
\end{equation}
Let $\mathbf{p}(e)\in\mathbb{R}^{C}$ denote the backbone-predicted class probabilities for edge $e$ (i.e., $\mathbf{p}(e)=\hat{\mathbf{p}}(e)$ in Eq.~\ref{eq:prob}). 
We capture prediction-level inconsistency by comparing the center prediction with its weighted neighborhood, including the neighborhood mean, the center-to-neighborhood deviation, and their weighted absolute differences.

For structural cues, we define an edge-level degree statistic vector $\mathbf{d}(e)$ that summarizes degree-related properties of $e$, such as the degree sum, degree difference, and degree product of its incident nodes.
We further define motif-level features $\mathbf{m}(e)$ using log-scaled counts of local motifs, including two-hop paths and triangles.

All signals are aggregated under the same neighbor-importance weights and concatenated to form the relative difference vector $\mathbf{r}_i$.
\begin{equation}
\mathbf{r}_i = \mathrm{concat}\Big(\mu_i(\mathbf{p}),\,\delta_i(\mathbf{p}),\,\mathbf{p}(e_i)-\mu_i(\mathbf{p}),\,
\mu_i(\mathbf{d}),\,\delta_i(\mathbf{d}),\,\mu_i(\mathbf{m}),\,\delta_i(\mathbf{m})\Big).
\end{equation}


\textbf{Relative-aware Nonconformity Score Construction.}  
Given the relative difference encoding $\mathbf{r}_i$ for edge $e_i$, we define the base nonconformity for assigning a candidate label $y\in\{0,1\}$ to $e_i$ as
\begin{equation}
a(e_i,y)=1-\hat{p}(y\mid e_i),
\end{equation}
where $\hat{p}(y\mid e_i)$ is the backbone-predicted probability for class $y$. We then incorporate neighborhood-relative discrepancies via
\begin{equation}
\tilde{a}(e_i,y) = a(e_i,y) + \lambda\, g(\mathbf{r}_i),
\label{eq:nonconformity_score}
\end{equation}
where $\lambda$ is a learnable scalar controlling the contribution of the relative component.

\subsection{Temporal Graph Calibration}  
Conformal calibration proceeds by estimating class-wise thresholds on a calibration split and then constructing prediction sets by thresholding nonconformity scores. However, in temporal transaction graphs, edge events are coupled through shared entities and evolve over time, leading to strong spatiotemporal dependencies. Directly calibrating on per-edge scores may thus be sensitive to local relational heterogeneity. To mitigate this issue, we adopt a diffusion-style temporal graph calibration step that smooths nonconformity over leakage-free temporal neighborhoods, while keeping the subsequent conformal quantile estimation unchanged. 

\textbf{Temporal Graph Neighborhood Score Diffusion.}
Given the intermediate nonconformity score $\tilde{a}_i$ for edge $e_i$ (Eq.~\ref{eq:nonconformity_score}) and its temporal neighborhood $\mathcal{N}_i$ constructed under the history-cutoff constraint ($t\le t_i$), We perform one-step diffusion within the prototype-conditioned subgraph $\mathcal{G}_i$ by smoothing $\tilde{a}$ over the neighborhood $\mathcal{N}_i$ using the prototype weights $\alpha_{ij}$:
\begin{equation}
a_i^{\text{tg}}
=
\beta\,\tilde{a}_i
+
(1-\beta)\sum_{e_j\in\mathcal{N}_i}\alpha_{ij}\tilde{a}_j.
\end{equation}
We set $\beta=0.5$ by default. This diffusion step yields a temporal-graph-aware nonconformity score $a_i^{\text{tg}}$ that incorporates both the center edge and its (weighted) historical context, and is computed without accessing future interactions.

\textbf{Class-conditional conformal thresholds.}
To address severe class imbalance, we use class-conditional (Mondrian) conformal calibration. Let $\mathcal{D}_{\text{cal}}$ denote the calibration set and $a^{\text{tg}}(e_i,y)$ the temporal-graph-aware nonconformity score for assigning label $y$ to $e_i$. For each class $y\in\{0,1\}$, let $\mathcal{D}_{\text{cal}}^{(y)}=\{(e_i,y_i)\in\mathcal{D}_{\text{cal}}: y_i=y\}$ and $n_y=|\mathcal{D}_{\text{cal}}^{(y)}|$. We compute the class-conditional threshold $q_y$ as an order statistic of the true-class scores:
\begin{equation}
q_y = a^{\text{tg}}_{y,(k_y)},\qquad k_y=\left\lceil (n_y+1)(1-\alpha)\right\rceil,
\end{equation}
where $a^{\text{tg}}_{y,(k_y)}$ denotes the $k_y$-th smallest value in the multiset $\{\, a^{\text{tg}}(e_i,y_i)\ :\ (e_i,y_i)\in\mathcal{D}_{\text{cal}}^{(y)} \,\}$.
Given $\{q_y\}$, we construct the prediction set for a test edge $e$ by including all labels whose nonconformity does not exceed the corresponding threshold:
\begin{equation}
\Gamma(e)=\{\, y\in\{0,1\}\ :\ a^{\text{tg}}(e,y)\le q_y \,\}.
\end{equation}


\textbf{Joint optimization. }
We jointly optimize the prototype module and the conformal objectives on the calibration split. For efficiency, we approximate the inclusion indicator with a sigmoid relaxation:
\begin{equation}
\tilde{\mathbb{I}}_{i,y}=\sigma\!\left(\frac{q_y-a^{\text{tg}}(e_i,y)}{\tau}\right),
\end{equation}
which approaches $\mathbb{I}(a^{\text{tg}}(e_i,y)\le q_y)$ as $\tau\to 0$. 
To discourage empirical miscoverage, we penalize threshold violations on the true-label nonconformity via a hinge surrogate:
\begin{equation}
\mathcal{L}_{\text{cov}}
=
\frac{1}{|\mathcal{D}_{\text{cal}}|}\sum_{(e_i,y_i)\in\mathcal{D}_{\text{cal}}}
\left[\frac{a^{\text{tg}}(e_i,y_i)-q_{y_i}}{\tau}\right]_+,
\end{equation}
where $[x]_+=\max(0,x)$. During backpropagation, we treat the empirical class-wise thresholds $\{q_y\}$ as constants.
Using $\tilde{\mathbb{I}}_{i,y}$, we define the surrogate efficiency loss as
\begin{equation}
\mathcal{L}_{\text{eff}}=\frac{1}{|\mathcal{D}_{\text{cal}}|}\sum_{e_i\in\mathcal{D}_{\text{cal}}}\sum_{y\in\{0,1\}}\tilde{\mathbb{I}}_{i,y},
\end{equation}
where $\mathcal{L}_{\text{eff}}$ is a smooth proxy of the expected prediction-set size. The overall training objective is
\begin{equation}
\mathcal{L}
=
\lambda_{\text{cov}}\,\mathcal{L}_{\text{cov}}
+
\lambda_{\text{eff}}\,\mathcal{L}_{\text{eff}}
+
\lambda_{\text{proto}}\,\mathcal{L}_{\text{proto}},
\end{equation}
where $\mathcal{L}_{\text{proto}}$ is the margin-based prototype loss defined in Sec.~\ref{sec:prototype}.

\section{Theoretical Analysis}
To explain the efficiency gains of \ourmethod, we analyze the empirical class-wise miscoverage gap. While standard split/Mondrian conformal coverage relies on exchangeability, temporal fraud graphs may involve relational dependence and temporal shift. We therefore decompose how calibration--test score-distribution shift contributes to the empirical gap, showing that prototype conditioning and temporal diffusion are designed to reduce this shift rather than establish a new finite-sample guarantee under arbitrary drift.

In our framework, each edge is additionally assigned to a prototype index $K\in\{1,\dots,K_0\}$ via the learned prototype bank (detailed in Section~4.1). Intuitively, prototypes partition the edges into behaviorally coherent groups, so that calibration and test edges within the same prototype share more similar score distributions.
Within class $y$, the empirical prototype weights and mode-conditional CDFs are
\begin{equation}
\begin{aligned}
\widehat\pi^{\mathrm{cal}}_{y}(k)&=\widehat{\Pr}(K=k\mid Y=y,\mathrm{cal}), \\
\widehat\pi^{\mathrm{test}}_{y}(k)&=\widehat{\Pr}(K=k\mid Y=y,\mathrm{test}),
\end{aligned}
\end{equation}
\begin{equation}
\begin{aligned}
\widehat F^{\mathrm{cal}}_{y,k}(t)&=\widehat{\Pr}(S^{\mathrm{cal}}\le t\mid Y=y,K=k,\mathrm{cal}), \\
\widehat F^{\mathrm{test}}_{y,k}(t)&=\widehat{\Pr}(S^{\mathrm{test}}\le t\mid Y=y,K=k,\mathrm{test}).
\end{aligned}
\end{equation}
Under severe class imbalance, a single global threshold can be dominated by the majority class. Class-conditional CP addresses this by computing a separate threshold for each class using only the true-class calibration scores:
\begin{equation}
\hat q_y := \widehat{\mathrm{Quant}}_{1-\alpha}\Big(\{S^{\mathrm{cal}}_{y,i}\}_{i=1}^{n_y}\Big).
\end{equation}
The empirical miscoverage on the test set and its deviation from the target level $\alpha$ are then
\begin{equation}
\widehat{\mathrm{mis}}_y :=
\frac{1}{m_y}\sum_{j=1}^{m_y}\mathbf{1}\{S^{\mathrm{test}}_{y,j}>\hat q_y\}, \qquad
\widehat{\Delta}_y := \big|\widehat{\mathrm{mis}}_y-\alpha\big|.
\end{equation}
To analyze when $\widehat{\Delta}_y$ becomes large, we decompose the miscoverage gap.
Since $\widehat{\mathrm{mis}}_y = 1-\widehat F^{\mathrm{test}}_y(\hat q_y)$, we have
\begin{equation}
\widehat{\mathrm{mis}}_y-\alpha
=
\underbrace{\Big((1-\alpha) -\widehat F^{ \mathrm{cal}}_y(\hat q_y)\Big)}_{\text{quantile discretization error}}
+
\underbrace{\Big(\widehat F^{\mathrm{cal}}_y(\hat q_y)-\widehat F^{\mathrm{test}}_y(\hat q_y)\Big)}_{\text{cal-test CDF shift}},
\end{equation}
where the first term reflects finite-sample discretization and is typically small, while the second term captures the distributional shift between calibration and test scores. We denote the latter as the empirical shift:
\begin{equation}
\label{eq:delta_y}
\big|\widehat{\delta}_y\big|
:=
\Big|\widehat F^{\mathrm{cal}}_y(\hat q_y)-\widehat F^{\mathrm{test}}_y(\hat q_y)\Big|.
\end{equation}
Reducing $|\widehat{\delta}_y|$ is the central goal of our method, as a smaller shift leads to tighter prediction sets at the target coverage.

Prototype conditioning can further reduce $|\widehat{\delta}_y|$. Since the class-conditional CDF decomposes as a mixture over prototypes,
\begin{equation}
\begin{aligned}
\widehat F^{\mathrm{cal}}_y(t)
&=\sum_{k=1}^{K_0}\widehat\pi^{\mathrm{cal}}_{y}(k)\,\widehat F^{\mathrm{cal}}_{y,k}(t),\\
\widehat F^{\mathrm{test}}_y(t)
&=\sum_{k=1}^{K_0}\widehat\pi^{\mathrm{test}}_{y}(k)\,\widehat F^{\mathrm{test}}_{y,k}(t),
\end{aligned}
\end{equation}
we obtain the following bound at the learned threshold $\hat q_y$:
\begin{equation}
\label{eq:proto_compress_bound}
\begin{aligned}
\big|\widehat{\delta}_y\big|
\le\;&
\underbrace{\sum_{k=1}^{K_0}\widehat\pi^{\mathrm{test}}_{y}(k)\,
\Big|\widehat F^{\mathrm{cal}}_{y,k}(\hat q_y)-\widehat F^{\mathrm{test}}_{y,k}(\hat q_y)\Big|}_{\text{within-prototype CDF discrepancy}}\\
&+\underbrace{\sum_{k=1}^{K_0}\Big|\widehat\pi^{\mathrm{cal}}_{y}(k)-\widehat\pi^{\mathrm{test}}_{y}(k)\Big|}_{\text{prototype-frequency mismatch}}.
\end{aligned}
\end{equation}
Prototype conditioning reduces $|\widehat{\delta}_y|$ through two mechanisms: (i) grouping behaviorally similar edges yields more homogeneous within-prototype score distributions, shrinking the first term; and (ii) learning stable prototypes that capture recurring patterns reduces the frequency mismatch in the second term.
After prototype conditioning, diffusion further stabilizes $s^{\mathrm{final}}$ against local spatiotemporal noise by smoothing each edge's score over its temporal neighborhood, empirically reducing
$|\widehat{\delta}_y|$ in Eq.~\eqref{eq:delta_y}.
\section{Experiment}\label{sec:exp}
\subsection{Experimental Setup}

\textbf{Datasets.} We evaluate \ourmethod on four datasets: the YelpChi dataset~\cite{rayana2015collective}, the small version of the Financial Fraud Semi-supervised Dataset (S-FFSD)~\cite{xiang2023semi}, the Financial Transaction Fraud Detection dataset (FTFD)~\footnote{\url{https://www.kaggle.com/datasets/aryan208/financial-transactions-dataset-for-fraud-detection}}, and the BankSim dataset~\cite{lopez2014banksim}. Detailed dataset introduction is listed in Appendix~\ref{sec:dataintro}

\textbf{Experimental Setup.} For the main experiments, we use TGN~\cite{rossi2020temporal} as the temporal graph backbone, pre-train it on the training set, and then freeze its parameters for calibration and conformal evaluation following the standard split-conformal protocol. To examine backbone transferability, we additionally evaluate \ourmethod with TGAT~\cite{xu2020inductive} on representative datasets. We use temporal splits to prevent future information leakage, with the calibration set taken from the time window before the test period. Backbone performance and TGAT results are reported in Appendix~\ref{sec:backbone}.
For baselines that do not natively support edge-level classification, we use a unified minimal adaptation by concatenating the two incident-node representations as the edge representation. For TGN-based models, edge representations and prediction scores are obtained from the interaction encoder conditioned on event history. Unless otherwise specified, edge neighbors are edges sharing a common node. Further setup details are provided in Appendix~\ref{sec:setup}.

\textbf{Evaluation Metric.} We evaluate all methods using \emph{coverage}, the fraction of test edges whose true label is included in the prediction set, and \emph{efficiency}, the average prediction set size (lower is better). Formal definitions are provided in Appendix~\ref{sec:metric}.

\subsection{Experiment Results}
\textbf{Conformal Prediction Performance.}
\MainResultTable
We compare \ourmethod with traditional conformal prediction methods (TPS~\cite{sadinle2019least}, APS~\cite{romano2020classification}, and RAPS~\cite{angelopoulos2020uncertainty}), graph-based conformal prediction methods (DAPS~\cite{zargarbashi2023conformal} and CF-GNN~\cite{huang2023uncertainty}), and non-exchangeability-based approaches (NCPNet~\cite{wang2025non}, NEX~\cite{barber2023conformal}, and NAPS~\cite{clarkson2023distribution}).

As shown in Table~\ref{tab:main_result}, \ourmethod consistently achieves the target coverage of 0.97--0.99 across all datasets, and yields the smallest prediction set sizes on S-FFSD, FTFD, and BankSim, with competitive efficiency on YelpChi. Traditional conformal methods often fail to reach the desired coverage because they ignore spatiotemporal dependencies among samples. Stronger baselines such as CF-GNN and NCPNet achieve valid coverage on most datasets, but produce larger prediction sets. For example, on S-FFSD, NCPNet and CF-GNN reach set sizes of 1.30 and 1.29, while \ourmethod achieves 1.16. On FTFD and BankSim, NCPNet produces set sizes of 1.15 and 1.14, compared with 1.07 and 1.04 by \ourmethod. This suggests that existing graph conformal predictors can become overly conservative under deceptive neighborhoods, which inflate nonconformity scores and calibrated thresholds.

To validate coverage and effectiveness on both fraud and benign classes, we further report class-wise coverage and efficiency in Table~\ref{tab:sep_main_result}.

\SepMainTable

From the class-wise perspective, most baselines under-cover the fraud class while over-covering the benign class. For instance, CF-GNN and NCPNet only reach fraud-class coverage of 0.82--0.90 and 0.89--0.93 across the four datasets, all below the 0.95 target, while their benign-class coverage consistently meets or exceeds it. This pattern reflects the limitation of a single global calibration threshold under severe imbalance, where the majority class dominates calibration. In contrast, \ourmethod attains target coverage for both classes, achieving 0.95--0.98 on fraud and 0.97--0.99 on benign edges. This indicates that prototype-conditioned subgraph calibration and relative-difference scoring mitigate noisy or uninformative neighborhoods and provide more reliable uncertainty control under extreme imbalance.

\textbf{Effect of Label-Conditioned Calibration.} Standard conformal baselines typically employ a global calibration threshold. In highly imbalanced fraud detection, global calibration can preserve marginal coverage yet exhibit uneven class-wise coverage, especially on the fraud class. We therefore also report label-conditioned variants for score-based baselines when applicable, which calibrate class-specific quantile thresholds using label-stratified calibration scores.
\LabelCondTable

As shown in Table~\ref{tab:lc_effect}, applying label-conditioned calibration consistently improves the fraud-class coverage compared with global calibration, alleviating the under-coverage issue under severe class imbalance. For example, TPS improves its fraud-class coverage from 0.90 to 0.95, and NCPNet from 0.93 to 0.95 after applying label-conditioned calibration. This indicates that a single global threshold can be dominated by the majority class, whereas class-wise thresholds better align the calibrated uncertainty with minority-class risk.

However, this gain often comes with a trade-off: while label-conditioned calibration improves fraud-class coverage, it leads to more conservative prediction sets, resulting in inflated set sizes. Specifically, TPS~+LC increases the fraud-class set size from 1.75 to 1.95, and CF-GNN~+LC from 1.47 to 1.62. In contrast, \ourmethod achieves strong fraud-class coverage of 0.96 with a set size of only 1.26, substantially lower than the best label-conditioned baseline NCPNet~+LC at 1.53, demonstrating improved efficiency under the same validity requirement.




\textbf{Ablation. }As shown at the bottom of Table~\ref{tab:main_result}, we ablate two key components in \ourmethod, i.e., the prototype-based subgraph extraction module (w/o Prot) and the relative-differential capture module (w/o Diff), to quantify their impact on conformal validity and efficiency.

Removing the prototype module consistently enlarges prediction sets while maintaining target-level coverage, suggesting that prototype-conditioned calibration mainly reduces the conservativeness of conformal thresholds. Compared with \ourmethod, w/o Prot increases the average set size by +0.09 (YelpChi), +0.11 (S-FFSD), +0.17 (FTFD), and +0.15 (BankSim), consistent with our motivation that prototypes filter out benign, calibration-irrelevant neighbors and yield more homogeneous calibration residuals, leading to tighter sets. Ablating the differential-capture module yields a smaller but consistent efficiency drop, increasing the set size by +0.14/+0.03/+0.11/+0.08 on YelpChi/S-FFSD/FTFD/BankSim. This indicates that neighborhood-relative discrepancies provide complementary signals beyond absolute backbone confidence, making nonconformity scores more informative and the prediction set size tighter.

\begin{figure*}[t]
    \centering
    \includegraphics[width=1\linewidth]{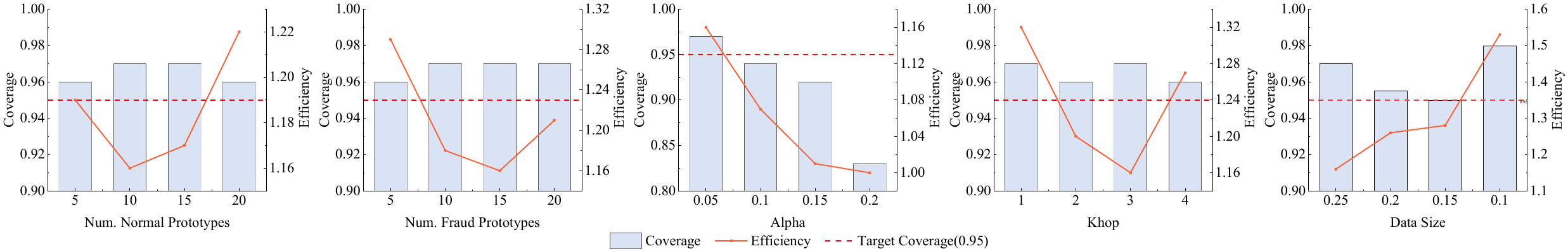}
    \caption{Visualization of sensitivity analysis under different hyperparameters and data size}
    \label{fig:sen}
    \vspace{-3mm}
\end{figure*}
\textbf{Parameter Sensitivity Analysis. }
To analyze the effect of the number of prototypes, we conduct a sensitivity study on the S-FFSD dataset, with results reported in Figure~\ref{fig:sen}. Increasing the number of prototypes first improves both coverage and efficiency, e.g. efficiency improves from 1.19 to 1.16 as normal prototypes increase from 5 to 10, but over-partitioning the embedding space yields noisy prototype assignments that destabilize calibration and enlarge prediction sets. The best-performing setting uses 15 fraud prototypes and 10 normal prototypes, reflecting the greater behavioral heterogeneity of fraudulent interactions. For the miscoverage level $\alpha$, coverage degrades gracefully as $\alpha$ increases from 0.05 to 0.1 while efficiency improves from 1.16 to 1.07, but at $\alpha=0.2$ coverage drops to 0.82. For the neighborhood hop $k$, a 3-hop neighborhood achieves the best coverage of 0.97 with best efficiency of 1.16; larger hops introduce distant, less relevant edges that inflate set sizes to 1.27 at $k=4$ without improving coverage. We also investigate the impact of calibration set size: as the calibration ratio decreases, coverage gradually drops and set sizes increase, where coverage inflates to 0.98 and efficiency degrades to 1.53, as limited calibration data undermine the prototype-based mechanism and the model degenerates toward standard temporal graph diffusion.

\subsection{Case Study and Visualization. }
\begin{figure}[t]
    \centering
    \includegraphics[width=1\linewidth]{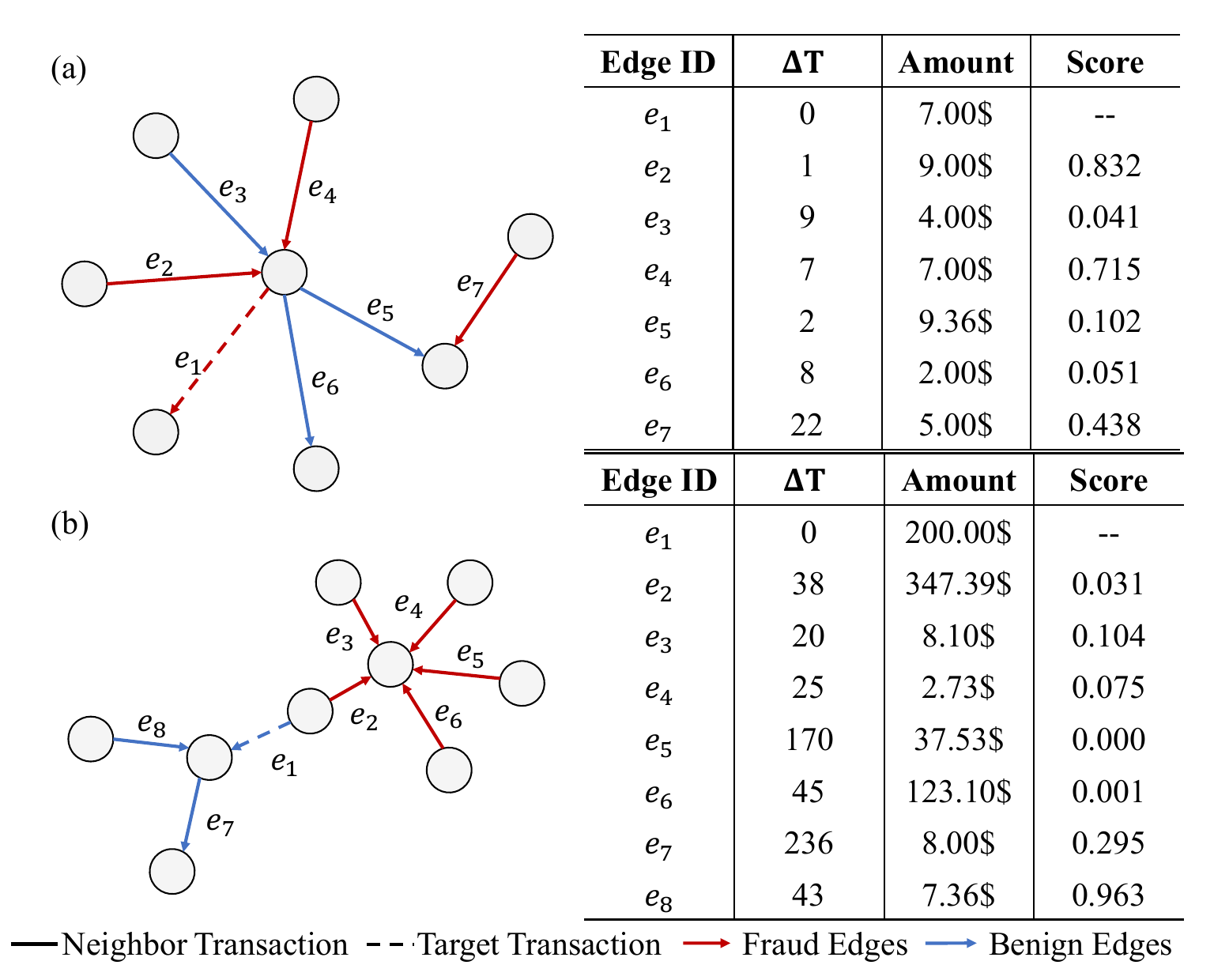}
    \vspace{-6mm}
    \caption{Visualization of temporal neighborhood subgraphs for a fraud (a) and a benign (b) target edge. Tables report $\Delta T$, amount, and the neighbor-importance weight $\alpha_{ij}$ (shown as \texttt{Score}). }
    \label{fig:case2}
    \vspace{-3mm}
\end{figure}

\begin{figure}[t]
    \centering
    \includegraphics[width=\linewidth]{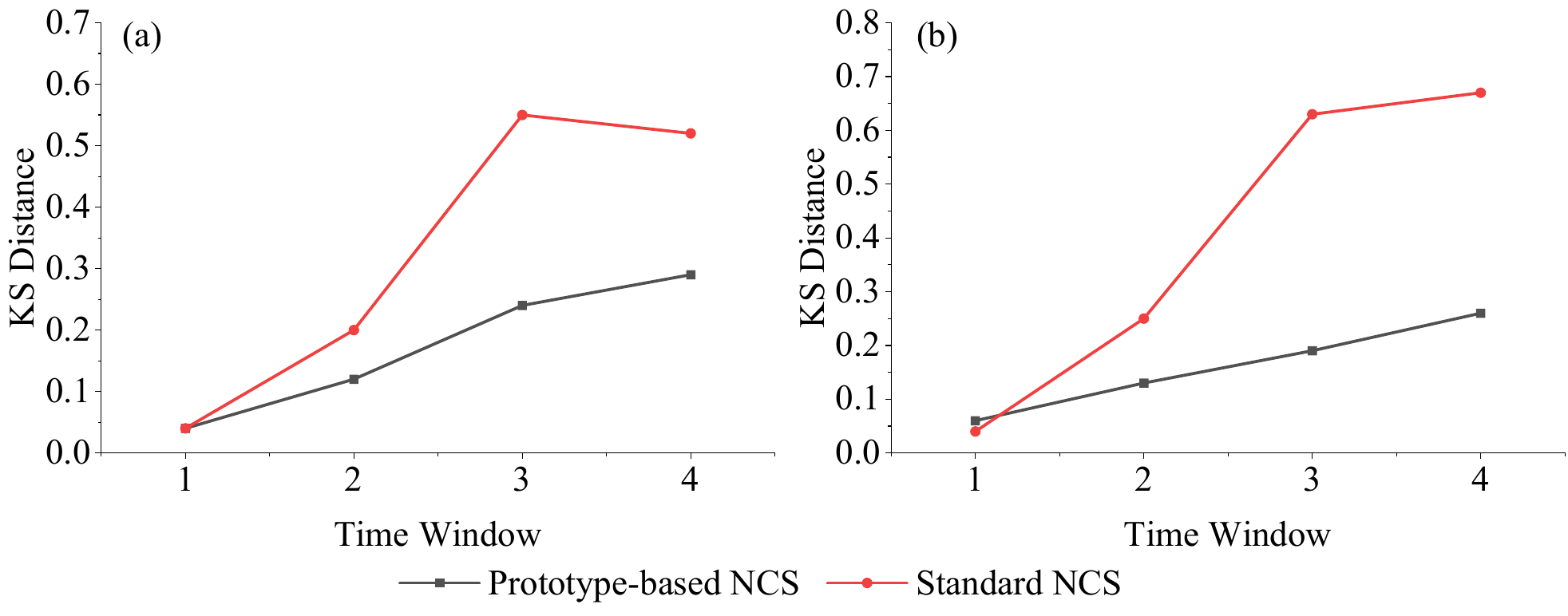}    \vspace{-6mm}
    \caption{Visualization of KS distance of Nonconformity Score(NCS) distributions over temporal windows on S-FFSD. (a) Benign class. (b) Fraud class.}
    \label{fig:ks_distance}
    \vspace{-3mm}

\end{figure}

To understand why prototype-based conformal calibration yields tighter prediction sets in fraud detection, we provide two complementary diagnostics on the S-FFSD dataset: (i) temporal neighborhood subgraphs showing how $\alpha_{ij}$ emphasizes calibration-relevant context (Figure~\ref{fig:case2}); and (ii) temporal-window analysis using KS distances between calibration and test NCS distributions (Figure~\ref{fig:ks_distance}) together with fraud-class coverage and efficiency (Table~\ref{tab:temporal_window_drift}).
As shown in Figure~\ref{fig:case2}, $\alpha_{ij}$ assigns higher importance to neighbors that provide a calibration-relevant context. For the fraud case (Figure~\ref{fig:case2}(a)), fraud-related neighbors generally receive larger $\alpha_{ij}$ than benign ones, and even a structurally more distant 2-hop fraud neighbor ($\Delta T=22$) can be up-weighted, reflecting prototype alignment beyond mere proximity. For the benign case (Figure~\ref{fig:case2}(b)), $\alpha_{ij}$ concentrates on high-score benign neighbors while down-weighting fraud-related distractors; meanwhile, a temporally distant benign neighbor ($\Delta T=236$) is assigned a smaller $\alpha_{ij}$ than closer benign neighbors despite its high score, suggesting that temporal proximity still modulates the weight among same-class edges.

To validate that prototype-based nonconformity scores mitigate the coverage gap under temporal drift, we split the test set into four temporal windows and compare each window with the calibration set using the KS distance of class-wise NCS distributions. As shown in Figure~\ref{fig:ks_distance}, the KS distance of the standard NCS increases monotonically over time, especially for the fraud class, while the prototype-based NCS remains consistently smaller and grows more slowly. This indicates that prototype-conditioned scoring reduces calibration--test score-distribution shift across time.

Table~\ref{tab:temporal_window_drift} further shows that this reduced shift leads to more stable class-conditional calibration under temporal drift. For NCPNet, fraud-class coverage drops from 0.98 to 0.92 across temporal windows, while fraud-class efficiency degrades from 1.42 to 1.66. In contrast, \ourmethod maintains fraud-class coverage from 0.97 to 0.95, with efficiency increasing only from 1.22 to 1.29. These results provide direct evidence that \ourmethod improves temporal robustness by stabilizing the fraud-class nonconformity distribution and maintaining tighter prediction sets over time.

\begin{table}[t]
    \centering
    \small
    \caption{Fraud-class coverage and efficiency across temporal test windows on S-FFSD. Lower efficiency indicates smaller prediction sets.}
    \label{tab:temporal_window_drift}
    \vspace{-1mm}
    \resizebox{\linewidth}{!}{
    \begin{tabular}{c cc cc}
        \toprule
        \multirow{2}{*}{Window} 
        & \multicolumn{2}{c}{NCPNet} 
        & \multicolumn{2}{c}{\ourmethod} \\
        \cmidrule(lr){2-3} \cmidrule(lr){4-5}
        & Coverage & Efficiency 
        & Coverage & Efficiency \\
        \midrule
        1 & 0.98 & 1.42 & 0.97 & 1.22 \\
        2 & 0.98 & 1.45 & 0.97 & 1.24 \\
        3 & 0.95 & 1.57 & 0.96 & 1.27 \\
        4 & 0.92 & 1.66 & 0.95 & 1.29 \\
        \bottomrule
    \end{tabular}
    }
    \vspace{-3mm}
\end{table}
\textbf{Computational Cost.}
We further analyze the computational cost introduced by \ourmethod beyond the frozen temporal GNN backbone. Let $M$ be the number of edges in the current split/window, $V$ the number of nodes, $\Delta$ the maximum historical degree, $U$ the number of unique edges in a calibration batch, $P$ the number of center-neighbor edge pairs, $C$ the number of classes, $K$ the number of prototypes, and $d$ the embedding dimension. The one-time preprocessing stage constructs history-only temporal neighborhoods and precomputes structural/motif features, requiring $O(M\Delta)$ time and $O(M+V)$ space. During calibration, motif aggregation uses precomputed two-hop and triangle features with $O(U+P)$ cost, prototype matching costs $O(UKd)$, and relative structural encoding costs $O(PC+U+P)$ per batch. Since $K$ is small by default ($15$ fraud and $10$ benign prototypes), the main overhead comes from temporal neighborhood construction and pairwise relative encoding.

Table~\ref{tab:runtime_memory} reports runtime and memory on S-FFSD. \ourmethod is more expensive than CF-GNN and NCPNet because of prototype-conditioned subgraph extraction and relative structural encoding, but the overhead remains moderate: preprocessing takes $79.43$ seconds with $1.88$ GiB peak memory, while calibration takes $0.947$ seconds per epoch with $2.18$ GiB peak GPU memory.

\begin{table}[h]
    \centering
    \small
    \caption{Runtime and memory comparison on S-FFSD. N/A indicates that the baseline does not require the corresponding preprocessing stage.}
    \label{tab:runtime_memory}
    \vspace{-1mm}
    \resizebox{\linewidth}{!}{
    \begin{tabular}{lcccc}
        \toprule
        Method 
        & Pre. Time 
        & Pre. Mem. 
        & Cal. Mem. 
        & Cal. Time \\
        & (s) & (GiB) & (GiB) & (s/epoch) \\
        \midrule
        CF-GNN & N/A & N/A & 0.85 & 0.315 \\
        NCPNet & N/A & N/A & 0.33 & 0.641 \\
        \ourmethod & 79.43 & 1.88 & 2.18 & 0.947 \\
        \bottomrule
    \end{tabular}
    }
    \vspace{-3mm}
\end{table}
\section{Conclusion}
We study conformal prediction for fraud detection on temporal interaction graphs, where benign-dominated neighborhoods and extreme class imbalance degrade calibration efficiency. We propose \ourmethod, a prototype-based conformal prediction framework that constructs fraud-informative calibration subgraphs via learned prototypes and stabilizes calibration with neighborhood-relative nonconformity scores and temporal diffusion. Experiments on four fraud datasets show that \ourmethod achieves the target coverage with consistently tighter prediction sets than conformal baselines, with particularly improved class-wise validity on the fraud class. 

\begin{acks}
The authors thank the anonymous reviewers for their constructive feedback and helpful suggestions.

\end{acks}
\bibliographystyle{ACM-Reference-Format}
\bibliography{ref}
\appendix
\section{Backbone and Dataset Robustness}\label{sec:backbone}

\textbf{Backbone performance on main datasets.}
To ensure that the backbone model (TGN) provides a reliable foundation for our framework, we report accuracy and F1 score across the four main datasets in Table~\ref{tab:backbone}.

\begin{table}[h]
\caption{Performance of the backbone model (TGN) across four fraud detection datasets in terms of Accuracy and F1 Score.}
\centering
\resizebox{\columnwidth}{!}{
\begin{tabular}{c|cc|cc|cc|cc}
\toprule
\multirow{2}{*}{Metric}
 & \multicolumn{2}{c|}{YelpChi}
 & \multicolumn{2}{c|}{S-FFSD}
 & \multicolumn{2}{c|}{FTFD}
 & \multicolumn{2}{c}{BankSim} \\
\cmidrule(lr){2-3} \cmidrule(lr){4-5} \cmidrule(lr){6-7} \cmidrule(lr){8-9}
 & Accuracy & F1
 & Accuracy & F1
 & Accuracy & F1
 & Accuracy & F1 \\
\midrule
TGN
 & 79.75
 & 65.57
 & 89.43
 & 77.82
 & 98.15
 & 90.32
 & 96.74
 & 84.48 \\
\bottomrule
\end{tabular}}
\label{tab:backbone}
\end{table}

\textbf{Backbone transfer with TGAT.}
To examine whether \ourmethod is tied to a specific temporal encoder, we additionally evaluate \ourmethod with TGAT~\cite{xu2020inductive} as an alternative backbone on S-FFSD and YelpChi. The same temporal split and class-conditional calibration protocol are used as in the main experiments. Table~\ref{tab:tgat_backbone} reports the TGAT backbone performance, and Table~\ref{tab:tgat_fraud} reports fraud-class coverage and efficiency against NCPNet. The results show that \ourmethod remains effective with TGAT, achieving higher fraud-class coverage and smaller prediction sets than NCPNet on both datasets.

\begin{table}[h]
\caption{Performance of the TGAT backbone on S-FFSD and YelpChi.}
\centering
\small
\begin{tabular}{lcc}
\toprule
Dataset & Accuracy & F1 Score \\
\midrule
S-FFSD & 89.64 & 78.02 \\
YelpChi & 81.43 & 66.32 \\
\bottomrule
\end{tabular}
\label{tab:tgat_backbone}
\end{table}

\begin{table}[h]
\caption{Fraud-class coverage and efficiency with TGAT backbone. Lower efficiency indicates smaller prediction sets.}
\centering
\small
\resizebox{\columnwidth}{!}{
\begin{tabular}{lcccc}
\toprule
\multirow{2}{*}{Method}
& \multicolumn{2}{c}{S-FFSD}
& \multicolumn{2}{c}{YelpChi} \\
\cmidrule(lr){2-3}\cmidrule(lr){4-5}
& Coverage & Efficiency
& Coverage & Efficiency \\
\midrule
NCPNet & 0.95 & 1.47 & 0.95 & 1.42 \\
\ourmethod & 0.96 & 1.31 & 0.96 & 1.35 \\
\bottomrule
\end{tabular}}
\label{tab:tgat_fraud}
\end{table}

\textbf{Supplementary AML benchmark.}
To further assess \ourmethod on a transaction-style AML benchmark, we include the AML Laundering Dataset (HI-SMALL)~\cite{altman2023realistic} as a supplementary dataset. HI-SMALL is better aligned with our temporal edge-level transaction setting than some commonly used alternatives. For example, Elliptic is organized into discrete time steps rather than a continuous interaction stream, while DGraphFin uses node-level labels rather than edge-level transaction labels. Table~\ref{tab:hismall_backbone} reports the backbone performance, and Table~\ref{tab:hismall_results} reports class-wise conformal results on HI-SMALL.

\begin{table}[h]
\caption{Backbone performance on HI-SMALL.}
\centering
\small
\begin{tabular}{lcc}
\toprule
Dataset & Accuracy & F1 Score \\
\midrule
HI-SMALL & 99.81 & 75.78 \\
\bottomrule
\end{tabular}
\label{tab:hismall_backbone}
\end{table}

\begin{table}[h]
\caption{Class-wise coverage and efficiency on HI-SMALL. Lower efficiency indicates smaller prediction sets.}
\centering
\small
\resizebox{\columnwidth}{!}{
\begin{tabular}{lcccc}
\toprule
Method & Fraud Cov. & Fraud Eff. & Benign Cov. & Benign Eff. \\
\midrule
TPS    & 0.93 & 1.92 & 0.95 & 1.13 \\
APS    & 0.95 & 1.96 & 0.95 & 1.23 \\
RAPS   & 0.92 & 1.94 & 0.96 & 1.19 \\
DAPS   & 0.93 & 1.56 & 0.95 & 1.09 \\
CF-GNN & 0.94 & 1.73 & 0.96 & 1.09 \\
NCPNet & 0.95 & 1.52 & 0.98 & 1.07 \\
NEX    & 0.92 & 1.75 & 0.97 & 1.10 \\
NAPS   & 0.95 & 1.71 & 0.96 & 1.08 \\
\ourmethod & 0.96 & 1.46 & 0.97 & 1.07 \\
\bottomrule
\end{tabular}}
\label{tab:hismall_results}
\end{table}

\section{Additional Component Ablation}\label{sec:additional_ablation}

\textbf{One-step temporal diffusion ablation.}
The main ablation in Table~1 evaluates the effect of removing the prototype module and the relative-difference scoring module. To further isolate the contribution of temporal score diffusion, we remove only the one-step history-only diffusion step while keeping prototype-conditioned scoring unchanged. As shown in Table~\ref{tab:onestep_ablation}, removing this step reduces fraud-class coverage and increases fraud-class prediction set size on both S-FFSD and YelpChi, indicating that the diffusion step provides complementary stabilization for class-conditional calibration under temporal dependence.

\begin{table}[h]
\caption{Ablation of the one-step temporal diffusion component. Lower efficiency indicates smaller prediction sets.}
\centering
\small
\resizebox{\columnwidth}{!}{
\begin{tabular}{lcccc}
\toprule
\multirow{2}{*}{Variant}
& \multicolumn{2}{c}{S-FFSD}
& \multicolumn{2}{c}{YelpChi} \\
\cmidrule(lr){2-3}\cmidrule(lr){4-5}
& Fraud Cov. & Fraud Eff.
& Fraud Cov. & Fraud Eff. \\
\midrule
w/o one-step diffusion & 0.93 & 1.35 & 0.95 & 1.61 \\
\ourmethod & 0.96 & 1.26 & 0.98 & 1.52 \\
\bottomrule
\end{tabular}}
\label{tab:onestep_ablation}
\end{table}

\section{Dataset Introduction}\label{sec:dataintro}
We provide detailed descriptions of the four fraud detection datasets used in our experiments. Summary statistics are shown in Table~\ref{tab:dataset}.
\begin{table}[h]
\caption{Summary statistics of the four fraud detection datasets.}
\centering
\resizebox{\columnwidth}{!}{
\begin{tabular}{l|cccc}
\toprule
Statistic & YelpChi & S-FFSD & FTFD & BankSim \\
\midrule
\# Samples        & 67,395     & 77,881     & 5,000,000     & 594,643     \\
\# Fraud (Positive) & 8,916   & 5,256     & 179,553     & 7,200     \\
\# Benign (Negative) & 58,479  & 24,387     & 4,820,447     & 587,443     \\
\# Unknown              & --       & 48,238  & --         & --      \\
Fraud Ratio (\%)   & 13.23\%     & 17.73\%     & 3.59\%     & 1.21\%     \\
Domain              & Review & Payment & Transaction & Transaction \\
\bottomrule
\end{tabular}}
\label{tab:dataset}
\end{table}

YelpChi~\cite{rayana2015collective} is a review dataset collected from Yelp.com for restaurants and hotels in the Chicago area, containing both recommended (benign) and filtered (spam/suspicious) reviews along with their associated users and businesses, and is widely used for review spam detection experiments.

S-FFSD~\cite{xiang2023semi} is a simulated, small-scale version of the Financial Fraud Semi-Supervised Dataset, consisting of transaction records with partially observed fraud labels and widely used for financial fraud detection.

FTFD~\footnote{\url{https://www.kaggle.com/datasets/aryan208/financial-transactions-dataset-for-fraud-detection}} dataset contains 5 million synthetically generated financial transactions designed to simulate real-world behavior for fraud detection research and machine learning applications.

BankSim~\cite{lopez2014banksim} is a synthetic bank payment transaction dataset generated by an agent-based simulator calibrated on aggregated statistics from a real Spanish bank, and it provides large-scale card payment records over a multi-month period with injected fraud behaviors and signatures, making it suitable for benchmarking and developing fraud detection methods without exposing any sensitive personal or proprietary transaction data.
\section{Evaluation Metric}\label{sec:metric}
To evaluate \ourmethod, we use two quality scores: \emph{coverage} and \emph{efficiency} (average set size) to report our performance. Given a test set $\mathcal{D}_{\text{test}}$ and prediction set $\Gamma(e_i)$ for each test edge $e_i$, we compute:
\begin{equation}
\mathrm{Coverage} \;=\; \frac{1}{|\mathcal{D}_{\text{test}}|}\sum_{(e_i,y_i)\in\mathcal{D}_{\text{test}}}\mathbb{I}\!\left[y_i \in \Gamma(e_i)\right],
\end{equation}
\begin{equation}
\mathrm{Efficiency} \;=\; \frac{1}{|\mathcal{D}_{\text{test}}|}\sum_{(e_i,y_i)\in\mathcal{D}_{\text{test}}}\left|\Gamma(e_i)\right|.
\end{equation}
When $\Gamma(e_i)=\emptyset$, we keep the coverage computation unchanged, but penalize efficiency by setting $|\Gamma(e_i)|=|\mathcal{Y}|$ (i.e., counting it as the full label set).

\section{Detailed Experiment Setup}\label{sec:setup}

\textbf{Data Splitting.}
For all datasets, we adopt a temporal split to respect the temporal ordering of interactions. Specifically, the earliest 55\% of edges are used for backbone training, the next 25\% for calibration, and the final 20\% for testing. No future information is accessible during any stage.

\textbf{Backbone Training.}
We use TGN~\cite{rossi2020temporal} as the shared backbone across all methods. The backbone is trained with a binary cross-entropy loss using the Adam optimizer with a learning rate of $\{1\text{e-}3, 1\text{e-}4\}$ selected via validation. Training runs for up to 200 epochs with early stopping based on validation F1 (patience = 10). After training, backbone parameters are frozen for all subsequent calibration and evaluation stages.

\textbf{ProtoCP Configuration.}
For the prototype module, we initialize $M$ fraud prototypes and $N$ normal prototypes in the edge-embedding space, with default values $M=15$ and $N=10$ selected based on the sensitivity analysis in Section~\ref{sec:exp}. The margin $\gamma$ in the prototype loss is set to 1.0. The subgraph extraction uses a 3-hop temporal neighborhood with the history-cutoff constraint. The diffusion coefficient $\beta$ is set to 0.5 by default. The target miscoverage level is $\alpha=0.05$ unless otherwise specified.

To stabilize prototype learning, we adopt a staged loss weighting strategy. During the first 50 training epochs, we optimize the model using only the prototype loss. After epoch 50, we incorporate the coverage and efficiency losses with weights $\lambda_{\text{cov}}=1.0$ and $\lambda_{\text{eff}}=0.5$, while the weight of the prototype loss $\lambda_{\text{proto}}$ is gradually decayed from its initial value (0.1) over the remaining epochs.

\textbf{Computational Environment.}
All experiments are conducted on a single NVIDIA V100 GPU (32GB) with PyTorch 2.6. Each experiment is repeated 5 times with different random seeds, and we report the mean and standard deviation.
\end{document}